\documentclass[letterpaper, 10 pt, conference]{ieeeconf}

\IEEEoverridecommandlockouts 

\usepackage{amsmath,amsfonts}
\usepackage{booktabs}
\usepackage{bm}
\usepackage{cite}
\usepackage{color}
\usepackage{siunitx}
\usepackage{graphicx}
\usepackage[nolist]{acronym}

\newacro{vs}[VS]{Visual Servoing}
\newacro{ibvs}[IBVS]{Image-Based \ac{vs}}
\newacro{pbvs}[PBVS]{Position-Based \ac{vs}}
\newacro{dvs}[DVS]{Direct \ac{vs}}
\newacro{dof}[DoF]{Degrees of Freedom}
\newacro{nn}[NN]{artificial Neural Network}
\newacro{ml}[ML]{Machine Learning}
\newacro{ssl}[SSL]{Self-Supervised Learning}
\newacro{ooi}[OOI]{Object of Interest}
\newacro{ros}[ROS]{Robot Operating System}
\newacro{il}[IL]{Imitation Learning}
\newacro{dl}[DL]{Deep Learning}

\title{\LARGE \bf Visual Servoing with Geometrically Interpretable Neural Perception}

\author{Antonio Paolillo, Mirko Nava, Dario Piga, Alessandro Giusti%
\thanks{This work was granted by the European  Union’s  Horizon  2020  Research and Innovation  Programs under Grant No. 871743 through the 1-SWARM project, the Hasler foundation through the EViRCo project, and the Swiss National Science Foundation (SNSF) through the NCCR Robotics.}
\thanks{All the authors are with Dalle Molle Institute for Artificial Intelligence (IDSIA), USI-SUPSI, Lugano, Switzerland {\tt name.surname@idsia.ch}%
\linebreak
~
\linebreak
\vspace{-0.75mm}
\noindent\fbox{\footnotesize\begin{minipage}{0.99\textwidth}\copyright 2022 IEEE.  Personal use of this material is permitted. Permission from IEEE must be obtained for all other uses, in any current or future media, including reprinting/republishing this material for advertising or promotional purposes, creating new collective works, for resale or redistribution to servers or lists, or reuse of any copyrighted component of this work in other works.
\end{minipage}}
\vspace{-1.5cm}%
}%
}%

\newcommand{\todo}[1]{#1}

\begin{document}

\maketitle
\thispagestyle{empty}
\pagestyle{empty}

\begin{abstract}
An increasing number of \todo{nonspecialist} robotic users demand easy-to-use machines. In the context of visual servoing, the removal of explicit image processing is becoming a trend, allowing an easy application of this technique. This work presents a deep learning approach for solving the perception problem within the visual servoing scheme. An artificial neural network is trained using the supervision coming from the knowledge of the controller and the visual features motion model. In this way, it is possible to give a geometrical interpretation to the estimated visual features, which can be used in the analytical law of \todo{the} visual servoing. The approach keeps perception and control decoupled, conferring flexibility and interpretability \todo{on} the whole framework. Simulated and real experiments with a robotic manipulator validate our approach. 
\end{abstract}

\section{Introduction}\label{sec:introduction}

Recent advancements in the fields of robotics and AI are leading machines to higher levels of reliability and autonomy. 
Specific application domains, like domotics and Industry 4.0, are showing the big potential of these technologies.
Indeed, the ubiquity of robots chased for decades is quickly becoming reality, although it brings novel needs and challenges. 
In particular, an increasing number of nonspecialist users ask for easy-to-use robots and no programming duties. 
Even in technical domains, like industry, practitioners with little or no expertise in robotics wish for plug-and-play platforms.

\todo{Today's progress of} \ac{dl} permits a relatively straightforward application of \acp{nn}, which are known to handle well high dimensional data \todo{and} complex perception tasks~\cite{Goodfellow:book:2016}.
In this work, we focus on how to combine \ac{dl} and control techniques to solve the perception problem in the specific context of \ac{vs}.
The traditional \ac{vs} paradigm relies on the \todo{image processing} to extract \todo{a suitable control feedback}.
It has to be properly designed, implemented, and normally tailored to the desired task.
To increase the easiness of use, \todo{also} for \todo{nonspecialists}, a possible way is the complete removal of the explicit image processing block from the \ac{vs} scheme. 
\todo{Furthermore, it is desirable to have a modular perception block, possibly transferable to other visual controllers.}
To this end, we propose to train a \ac{nn} that \todo{derives} supervision from the knowledge of the control structure and the robot motion to provide neural feedback from monocular images.
The proposed framework \todo{keeps} the classic perception-and-control scheme where the feedback extraction algorithm is \todo{left} outside the control block.
Nevertheless, our perception model is made aware 
of the \ac{vs} structure and the visual features motion model,
leveraging this information \todo{to give a geometric interpretation to the neural feedback.}
This approach differs from an end-to-end where sensing and action are, instead, coupled.
We claim that keeping the \todo{controller} analytical structure \todo{enables} higher \todo{flexibility} of the entire framework, besides customization of the downstream control.

The rationale of relying on \ac{dl}, or other sorts of \ac{ml}, to avoid the explicit image processing in the \ac{vs} has already been proposed in different forms \todo{(see Sec.~\ref{sec:rw})}.
However, we speculate that neural perception, control, and modeling aspects can bring a beneficial synergy to provide the \ac{vs} scheme with effective neural feedback (Sec.~\ref{sec:method}).
Our approach is to train a neural perception model oriented to the \ac{vs} task so that the prediction provides a robust and tailored \todo{control feedback}.
Using the setup detailed in Sec.~\ref{sec:setup}, we validate our approach with simulations and real experiments (Sec.~\ref{sec:results}).
Final remarks in Sec.~\ref{sec:conclusion} conclude the paper.

\section{Related work}\label{sec:rw}

Classic \ac{vs} uses visual features extracted from camera images to control robots~\cite{Chaumette:ram:2006,Chaumette:ram:2007}.
In \ac{ibvs}, they are directly measured on the image, whereas \ac{pbvs} uses visually reconstructed camera poses.
The basic formulation has been expanded to planning~\cite{Mezouar:tro:2002,Chesi_tro:2007}, optimization-based control~\cite{Agravante:ral:2017,Paolillo:ral:2018,Mingo:icra:2021}, model predictive control~\cite{Allibert:tro:2010,Sauvee:cdc:2006,Paolillo:icra:2020} and integrated with \ac{ml}-based concepts~\cite{Paolillo:icra:2020,Paolillo:icra:2022}.
In any case, the images have to be explicitly processed to properly detect, track and match the visual features.

More advanced approaches avoid the explicit image processing considering the whole image as feedback.
This newer class of \ac{vs}, known as direct \ac{vs}, proposes to use, e.g., photometric moments~\cite{Bakthavatchalam:tro:2018}, luminance of the pixels~\cite{Collowet:tro:2011},  histograms~\cite{Bateux:ral:2017} or Gaussian mixtures~\cite{Crombez:tro:2019} as visual features.

Recently, \ac{ml} techniques have been considered to solve the image processing problem in \ac{vs}.
In~\cite{castelli2017machine} a Gaussian Mixture Model is used to imitate the classic VS behavior using a reduced \todo{image resolution}.
In~\cite{Jin:iros:2020}, \todo{task-relevant features are extracted by a \ac{nn} to build a more complex feedback and the visual error.} 
The idea of considering the whole image to avoid specific image processing is well suited to the application of \todo{\ac{dl}}, 
which efficiently handle dense visual information~\cite{Goodfellow:book:2016}.
Indeed, nowadays many works propose to learn the relative camera-scene pose from monocular images (e.g., see~\cite{Li:eccv:2018,Deng:icra:2020,Nava:2021:ral,Nava:ral:2022}), which in principle can be used to realize \ac{pbvs}.
Specifically to the VS case, several works rely on \ac{dl} to infer from images the relative camera displacement~\cite{Saxena:icra:2017,Yu:iros:2019,Bateux:icra:2018}.
In~\cite{Felton:ral:2022} visual features are extracted from an autoencoder latent space and their motion model computed, to be considered in the analytical \ac{vs} law.
Deep visual features and their dynamics are learned by an agent in~\cite{Lee:iclr:2017}.

In end-to-end approaches, a \ac{dl} architecture computes control commands directly from images, avoiding the explicit visual feature extraction by coupling the perception and control problem~\cite{Levine:jmlr:2016}.
End-to-end architectures use a \ac{nn} to first learn a representation of the image and another \ac{nn} to regress velocity commands~\cite{Puang:iros:2020,Felton:icra:2021}.
In~\cite{Mantegazza:icra:2019} an end-to-end is compared to a cascade of two \acp{nn} computing the target pose and the command for an aerial drone.

Our approach uses \ac{dl} to extract from monocular images geometrically-interpretable visual features needed to close the \ac{vs} loop.
In principle, it can learn any feedback described by the \ac{vs} law considered during the training \todo{phase,} being \todo{it} a Cartesian pose or an image feature.
It differs from an end-to-end approach, since we maintain the classic perception-control paradigm, keeping the controller structure untouched downstream the perception block.
Our neural perception derives supervision from the control knowledge so that it can provide \todo{a} feedback tailored to the \ac{vs}.
Furthermore, keeping the control operation outside the perception block allows major flexibility of the whole framework.
Being formally decoupled from the control\todo{,} the trained perception model can be deployed
within different control alternatives.

\section{Preliminaries}\label{sec:background}

In this section, we introduce the basic concepts and the required nomenclature to formalize our approach.
We recall the classic formalism adopted for 
\ac{vs} and \ac{nn}, and all the variables of interest that are involved in our discussion.

\subsection{Visual servoing}

In \ac{vs}, visual features $\bm{s}\in\mathbb{R}^f$ \todo{serve as} feedback, extracted or reconstructed from images, required to close the control loop realizing the desired robot behavior~\cite{Marchand:ras:2005,Chaumette:ram:2006,Corke:book:2017}.
In general, $\bm{s}$ depends on the camera calibration and/or Cartesian information~\cite{Chaumette:ram:2006,Corke:book:2017}; examples are points or lines on the image \todo{(in \ac{ibvs})} or reconstructed camera poses (in \ac{pbvs}).

Considering eye-in-hand configurations and a fixed-base robot with $n$ \ac{dof}, the time derivative of the visual features $\dot{\bm{s}}$ is directly related to the robot joints velocity $\dot{\bm{q}}\in\mathbb{R}^n$ through the relationship $\dot{\bm{s}} = \bm{J} \dot{\bm{q}}$, where $\bm{J} \in \mathbb{R}^{f\times n}$ is the Jacobian matrix.
Assuming a non-moving target and a constant reference, the classic VS law used to control the robot is $\dot{\bm{q}} = -\lambda \hat{\bm{J}}^+ (\bm{s} - \bm{s}^\ast)$~\cite{Chaumette:ram:2007, Marchand:ram:2005} where $\lambda$ is a positive control gain tuning the convergence rate; $\bm{s}^\ast \in\mathbb{R}^f$ are the desired visual features; $\hat{\bm{J}}^+ \in \mathbb{R}^{n\times f}$ is the pseudo-inverse of an approximation of the Jacobian.
At convergence, under specific assumptions~\cite{Chaumette:ram:2006}, the \ac{vs} law makes sure that the measured features match with their desired counterparts.
As a consequence, the camera is driven to the desired pose. 

The following considerations about the Jacobian matrix shall be mentioned.
It is composed as $\bm{J} = \bm{L}(\bm{s}) \,\bm{V} \bm{J}_r(\bm{q})$~\cite{Chaumette:ram:2007,Marchand:ram:2005}, where: $\bm{L} \in \mathbb{R}^{f \times 6}$ is the interaction matrix that depends on the features and relates their motion to the camera velocity; $\bm{V} \in \mathbb{R}^{6\times 6}$ is the constant matrix transforming the velocities from the end-effector to the camera frame; $\bm{J}_r \in \mathbb{R}^{6\times n}$ is the robot Jacobian in the end-effector frame and depends on the robot configuration $\bm{q}$.\footnote{If not explicitly mentioned, the dependency of the matrices on the robot configuration and the visual features is omitted for brevity.}
The Jacobian approximation is mainly due to the interaction matrix \todo{that} depends on generally unknown spatial information, e.g. the \todo{visual features} depth; an estimation or approximation is normally \todo{used,} see e.g., \cite{Chaumette:ram:2006,Allibert:tro:2010, Marchand:ram:2005}.
Furthermore, the Jacobian can become singular for particular robot or features configurations, with disruptive consequences on the correct \todo{control} convergence. 
A damping term can be added to the control law, paid at the cost of lower tracking performances~\cite{robotics:book}.

\subsection{Artificial neural networks}

We aim at extracting the \ac{vs} feedback using a \ac{nn}, tackling a pure perception task.
We use the term \emph{model} to denote the function $\bm{m}$ implemented by the \ac{nn} and predicting the visual features from an image, i.e. $\bm{s} \approx \bm{m}(\bm{i}|\bm{\theta})$, where $\bm{i}\in\mathbb{R}^{whc}$ is the vectorized content of the image with a resolution of $w\times h$ pixels and $c$ channels; $\bm{\theta}\in\mathbb{R}^m$ is the set of weights\footnote{In what follows, we omit the dependency of the models on the weights.} obtained optimizing a loss function ${\cal L}$ in the training phase.

The \todo{majority} of \ac{nn}-based methods rely on labels, i.e. the known value of the target variable contained in the data, to train models in a supervised fashion.
Nevertheless, especially in robotics, the labeling procedure 
is costly and demanding.
Alternative approaches use automated procedures to label the data, so that a bigger amount of information crucial for the performance of \ac{dl} techniques, are made available.
In this work, we propose to learn proper visual features for the \todo{\ac{vs}}, avoiding the processing required to explicitly label the images with visual features.
Instead, we extract supervision from the knowledge of the \ac{vs} and the motion model.

\section{Neural feedback for visual servoing}\label{sec:method}

The ultimate aim of the work is to derive a \ac{vs} law with the visual feedback provided by a \ac{nn}:
\begin{equation}
    \dot{\bm{q}} = - \lambda \hat{\bm{J}}^+ \Big( \bm{m}(\bm{i}) - \bm{m}(\bm{i}^\ast) \Big)
    \label{eq:control_embedded}
\end{equation}
where $\bm{i}^\ast$ indicates the desired image that the camera would see at the completion of the task. 

Our goal is to learn a neural perception model $\bm{m}$, such that the estimated visual features can be geometrically interpreted and thus used by a classic VS law.
The task is defined as follows: find proper visual features for the VS law without any explicit image processing, directly using the raw uncalibrated image captured by the robot on-board camera.
To this end, our model derives supervision directly from demonstrations of the VS, avoiding the need for explicit labels for the visual features.
Training data, collected from the robot itself while it realizes several executions of the visual task, is a sequence of the following tuple:
\begin{equation}
 \langle \, \dot{\bm{q}}_k, \bm{J}_{r,k}, \bm{i}_k \, \rangle,~k=1,\dots,N
 \label{eq:data}
\end{equation}
where $N$ is the number of samples.
Note that in the data there is no knowledge about the visual features.
All the required information is given by the robot standard sensory equipment, composed of joint encoders and a monocular camera, with no further processing and readily accessible.

\subsection{Supervising the perception task imitating the controller}\label{sec:loss_ci}

Proper visual features can be reconstructed by imitating the desired VS control behavior.
Thus, to train our model, we shall consider the following \emph{control imitation} loss function:
\begin{equation}
    {\cal L}_\text{CI} = \frac{1}{N} \sum_{k=1}^N \left\lVert \dot{\bm{q}}_{k} + \lambda \hat{\bm{J}}_k^+ \Big( \bm{m}(\bm{i}_k) - \bm{m}(\bm{i}_k^\ast) \Big) \right\rVert_1
    \label{eq:loss_control_imitation}
\end{equation}
that forces the model to learn visual features such that their use in the \ac{vs} law~\eqref{eq:control_embedded} imitates the demonstrated commands. 
We are not interested in learning the controller, as in  e.g. end-to-end approaches.
The knowledge of the control structure is instead leveraged to build a perception model tailored to the \ac{vs}. 
As a result, the visual features are learned without the need for their explicit knowledge in the training data.
In practice, the supervision of the learning process is provided by the imitation of the control behavior.
Note that in~\eqref{eq:loss_control_imitation} the reference image $\bm{i}_k^\ast$ is taken as the image captured at completion of each demonstration.

Recalling Sec.~\ref{sec:background}, the Jacobian in~\eqref{eq:loss_control_imitation} has this shape:
\begin{equation}
    \hat{\bm{J}}_k = \hat{\bm{L}}_k \big( \bm{m}(\bm{i}_k)\big) \, \bm{V} \, \bm{J}_{r,k} \big( \bm{q}_k \big)
\end{equation}
where $\bm{J}_{r,k}$ depends on the robot configuration and its numerical value is taken from the data~\eqref{eq:data}, while $\bm{V}$ is constant and known in advance.
Instead, the interaction matrix $\bm{L}_k$ depends on the visual features, i.e. the prediction of the model.
It is worth recalling that the interaction matrix also depends on the features' depth; their known value at the target is used as an approximation.
Furthermore, the analytical structure of the interaction matrix depends on the number and the kind of visual features, which are decided in advance in the training procedure (see Sec.~\ref{sec:setup}). 
This design choice allows us to format the structure of the interaction matrix, which helps the learning process to find a particular geometrical interpretation of the prediction.
Indeed, by imitating the control law, which has a precise structure grounded on the geometry of the visual features, it is possible to find in the data the correspondence between raw images and the geometrical interpretation requested by the \ac{vs} feedback.

During the training of the model, especially at the first epochs, the model might provide naive features configurations.
This issue possibly leads to a singularity of the Jacobian and a resultant ill-posed inversion problem. 
Thus, in~\eqref{eq:loss_control_imitation}, in place of $\hat{\bm{J}}_k^+$, it is considered:
\begin{equation}
    \left( \hat{\bm{J}}_k^\top \, \hat{\bm{J}}_k + \sigma^2 \bm{I}_n \right)^{-1} \hat{\bm{J}}_k^\top
\end{equation}
where $\sigma$ is a damping term used to better handle the inversion of the Jacobian~\cite{robotics:book}  and $\bm{I}_n$ is the $n\times n$ identity matrix.

\subsection{State consistency}\label{sec:loss_sc}

The temporal sequence of images and joint velocities is a rich source of supervision that can be exploited during the learning process.
In fact, assuming that the object of interest is static during the data collection, the evolution of the visual features is well described by its motion model: given the visual features' estimate $\bm{m}\left(\bm{i}_{k}\right)$, the joint velocities $\dot{\bm{q}}_k$ and the Jacobian $\bm{J}_k$ at timestep $k$, we expect the neural output to be geometrically consistent with the motion model, i.e. producing at timestep $k + 1$ the estimate $ \bm{m}\left(\bm{i}_{k}\right) + T \bm{J}_{k} \dot{\bm{q}}_k$.
We take advantage of this information and instantiate a state-consistency loss \todo{as in}~\cite{nava2021ral}, which penalizes models that have an eradic prediction, failing to be consistent with the scene:
\begin{equation}
    {\cal L}_\text{SC} = \frac{1}{N}\sum_{k=1}^{N-1} \left\lVert \bm{m}(\bm{i}_{k+1}) - \bm{m}\left(\bm{i}_{k}\right) - T \bm{J}_{k} \dot{\bm{q}}_k \right\rVert_1
    \label{eq:loss_geometrical_consistency}
\end{equation}
being $T$ the time difference between two consecutive timesteps.
The effect of this loss is to further enforce the desired geometrical interpretation of the neural perception. 

\subsection{Image reconstruction as pretext task}\label{sec:loss_ae}

A particular class of \acp{nn}, called auto-encoder, uses an encoder to reduce high-dimensional images into a small-size latent variable, and a decoder to reconstruct the input image from the latent variable.
The latent space is thus meant to be a highly informative domain where the most important notions of the image are condensed.
Therefore, it is reasonable to extract the feedback of our VS law~\eqref{eq:control_embedded} from the latent variable of an auto-encoder.
For this reason, we consider the \emph{auto-encoding} loss function:
\begin{equation}
    {\cal L}_\text{AE} = \frac{1}{N}\sum_{k=1}^N \left\Vert \bm{i}_k - \bm{\delta} \Big( \bm{\varepsilon}(\bm{i}_k) \Big) \right\Vert_1
    \label{eq:loss_autoencoder}
\end{equation}
where the function $\bm{\delta}$ and $\bm{\varepsilon}$ indicate the decoding and encoding parts, respectively.
Even though the latent variable is an effective representation of the image and conceptually close to the visual feature, its semantic interpretation is entrusted to the \ac{nn}.
To be correctly interpreted and actually used in the controller~\eqref{eq:control_embedded}, the visual features need to be extracted from the latent variable with an additional piece of network that we call \emph{head} and denote with $\bm{h}(\cdot)$. 
Therefore, the neural visual features have this form:
\begin{equation}
\bm{m}(\bm{i}_k) = \bm{h} \Big( \bm{\varepsilon}(\bm{i}_k) \Big).
\end{equation}
The head, trained by considering the loss functions of Sec.~\ref{sec:loss_ci} and Sec.~\ref{sec:loss_sc}, gives the correct interpretation to the output of the encoder.
Furthermore, it allows setting a desired dimension $f$ of the neural visual features.

Note that the auto-encoding can be seen as a \emph{pretext task}~\cite{jing2020self}, i.e. an auxiliary task that is of no direct interest, but whose solution stimulates the better performance of the \emph{end task}, in our case the perception task. 
Practically speaking, the pretext task helps to find patterns in the data that are useful for the solution of the end task.
Most importantly, as for~\eqref{eq:loss_control_imitation}, this strategy is particularly convenient because we can exploit unlabeled data and thus count on a bigger amount of information to solve our problem.

\subsection{Regularization term}

The loss functions described in the previous sections are enough to solve our perception task.
However, it has to be considered that the neural model is not enforced to produce an output easy to interpret for a human.
In fact, the \ac{nn} can find infinite solutions to our problem, all equally valid. 
For instance, it might produce point features outside the image plane borders and randomly displaced around the image of the object to track.
To overcome this issue, we add an heuristic-based regularization term ${\cal L}_\text{R}$ that solves the ambiguity of the \ac{nn} solutions by constraining particular geometrical properties.
For example, it can be used to force the desired prediction $\bm{m}(\bm{i}^\ast)$ to be visible on the image. 

\section{Experimental setup}\label{sec:setup}

To test our framework, we considered the $7$ \ac{dof} robotic manipulator by Franka Emika~\cite{franka}.
The robot is equipped at the end-effector with an Intel RealSense depth camera D435, used as a monocular camera, streaming images at the nominal frame-rate of $30$~Hz with a resolution of $640 \times 480$~pixels.
The robot also provides a measurement of the Jacobian and the received joint velocity commands at a frequency of $200$~Hz.
Thus, we had easy access to the required information to construct the dataset~\eqref{eq:data}.
The framework and the communication with the robot are implemented in Python within the \ac{ros}~\cite{ros} infrastructure; the \ac{nn} implementation is built upon the PyTorch library~\cite{Paszke:nips:2019}. 

\subsection{Data collection}

Data is collected within the robotic simulator Gazebo~\cite{Koenig:iros:2004}, 
where the exact knowledge of the robot and its working environment is available.
In Gazebo, the robot carries out multiple executions of a VS task, consisting in framing an \ac{ooi} at the center of the image captured by the onboard camera.
In principle, to collect proper joint velocity commands, the VS task could be demonstrated in different forms, e.g. by teleoperation, with a classic VS law, or using other control structures.
In our setup, the VS task is demonstrated by an \emph{ideal} classic \ac{ibvs} using 4 point visual features.
Thus, we consider the interaction matrix of point features in our learning procedure, and in the computation of the loss functions~\eqref{eq:loss_control_imitation} and~\eqref{eq:loss_geometrical_consistency}. 
As a consequence, the model will be forced to give the geometrical interpretation of points to its estimate.
\todo{In principle, different numbers and kinds of features may be considered, by selecting the corresponding interaction matrix in the training phase.}
The visual features are geometrically reconstructed from the 4 vertexes of the bounding box containing the \ac{ooi} and projecting them on the image using the camera projection model.
Thanks to the perfect knowledge of the camera model and the pose of the \ac{ooi} w.r.t. the camera frame, it is possible to realize the \ac{vs} task with high accuracy. 
Indeed, this demonstrator serves as a sort of \emph{oracle}, providing the learning procedure with ideal data.
However, the only collection of ideal situations might be not enough to guarantee the variability of data needed to reach generality performances during the deployment of the model.
In practice, with an ideal \ac{ibvs}, the smooth convergence produces very similar images, most of which have the \ac{ooi} at the center of the image.
To increase \todo{data} variability, we adopt a two-steps procedure: first, the robot camera is driven to a random pose, and then converges driven by the ideal \ac{ibvs} introduced above.
In both the steps, we collect the commands as computed by the \ac{ibvs}.
Furthermore, we adopt a domain randomization~\cite{tobin2017domain} strategy: color and texture of the background, lighting conditions, the planar pose of the \ac{ooi}, and the initial configuration of the robot are all randomized.
Overall, we collected in the simulator the equivalent of 2 hours of data, totaling to approximately 100k examples, generated from 500 task demonstrations, out of which 83.5k examples were from the training set, and 16.5k from the validation set.
Once collected, data is shuffled by pairs of two (to allow the implementation of the state-consistency loss, which needs at least two consecutive time samples) and organized in batches.

\subsection{Training}

Our model is trained for 100 epochs with gradient descent, using the combination of loss functions described in Sec.~\ref{sec:method}:
\begin{equation}
    {\cal L} = \lambda_\text{CI} {\cal L}_\text{CI} + \lambda_\text{AE} {\cal L}_\text{AE} + \lambda_\text{SC} {\cal L}_\text{SC} + \lambda_\text{R} {\cal L}_\text{R}
    \label{eq:loss}
\end{equation}
where the scalars $\lambda_\text{AE}$, $\lambda_\text{CA}$, $\lambda_\text{SC}$ and $\lambda_\text{R}$ are introduced to weigh the \todo{loss terms}, 
allowing the tuning of the different components.
In our setup, we heuristically set these parameters to one.
The optimizer of choice is Adam~\cite{adam} with a fixed learning rate of $1\mathrm{e}^{-4}$.
To synthetically increase the amount of training data, we apply data augmentation in the form of additive gaussian noise, random brightness, and contrast.

\subsection{\ac{nn} architecture}

We consider a convolutional \ac{nn} architecture composed of three main blocks: encoder, decoder, and head.
The encoder consists of a series of 4 convolutions with stride 2 followed by ReLU non-linearity, halving each time the size of the image, followed by the bottleneck of size 64.
The decoder consists of the bottleneck, followed by the same number of convolutions present in the encoder part, but with stride 1 and an up-sampling of the previous activation map, doubling each time the image size.
The head takes the latent representation from the bottleneck and through a series of 3 feed-forward layers with ReLU non-linearities, and the final layer with TanH non-linearity produces the
visual features. 
The model output is finally multiplied by a scaling factor, allowing the estimation to vary in a wider range and handle situations where the \ac{ooi} is partially outside the camera field of view. 

\section{Results}\label{sec:results}
\begin{figure}
    \centering
    \includegraphics[trim={0, 10, 0, 0}, clip,width=\columnwidth]{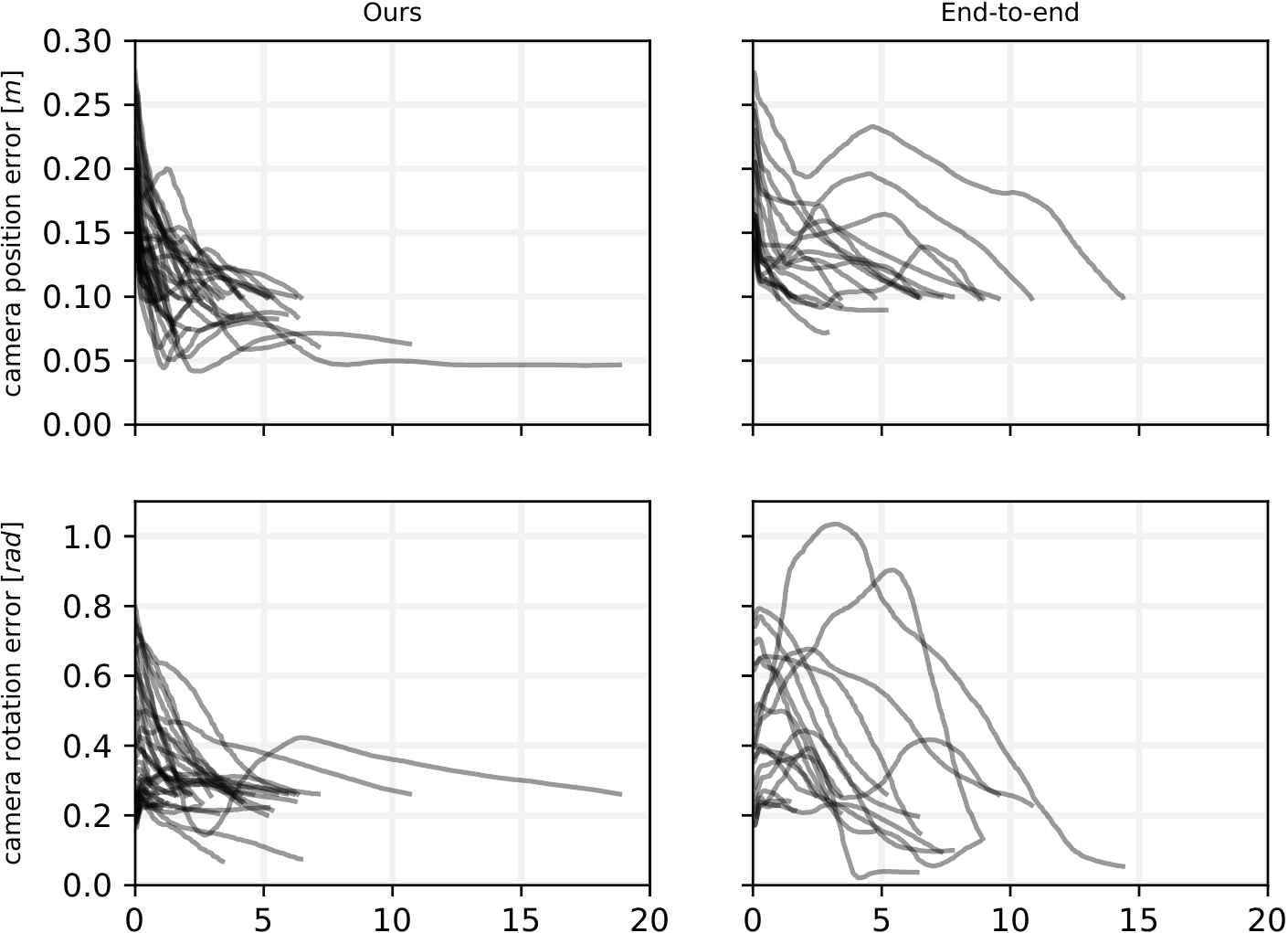}
    \\[10pt]
    \includegraphics[width=\columnwidth]{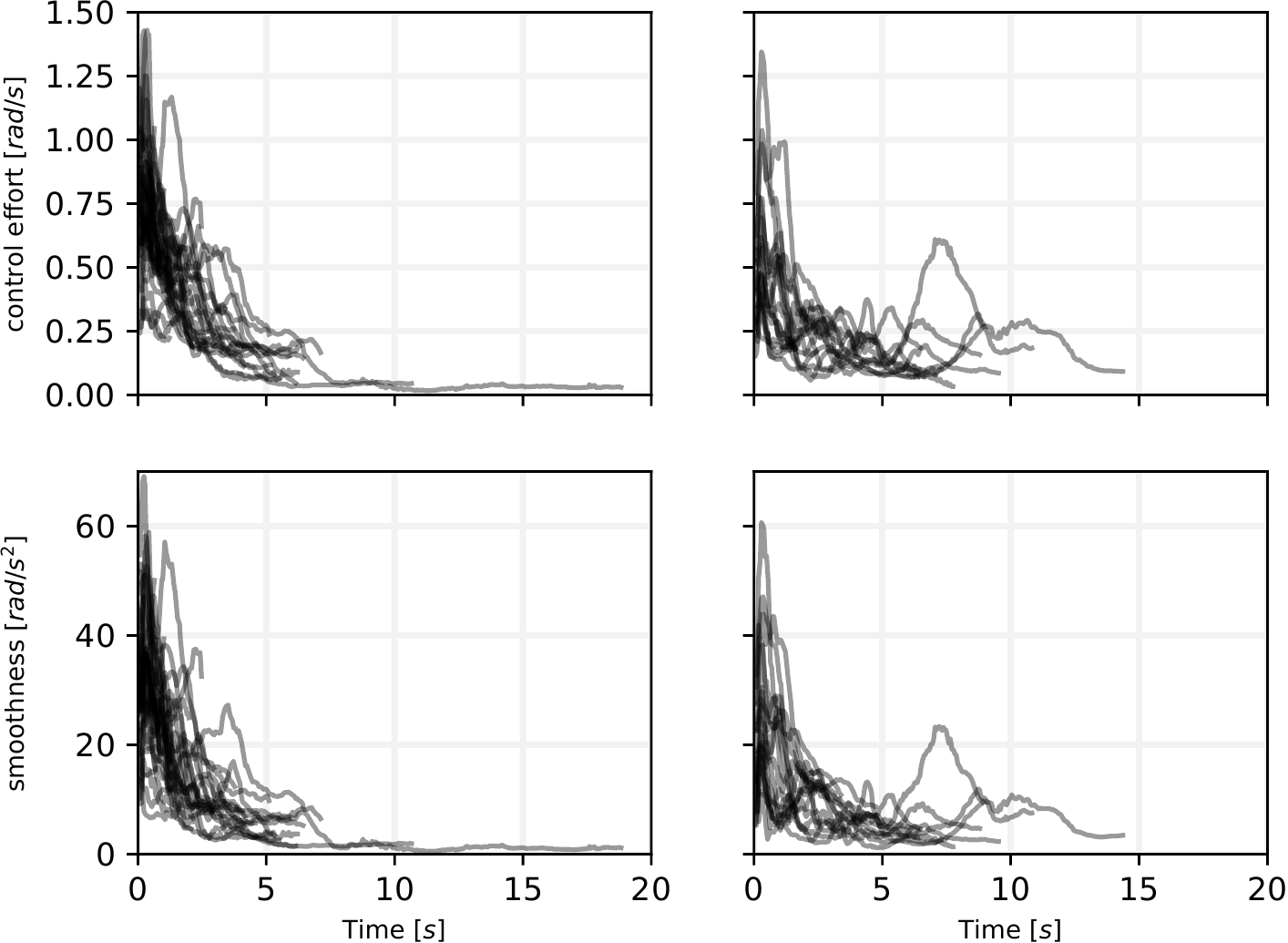}
    \caption{Control metrics evolution of the successful executions carried out by the analysis comparing our model with an end-to-end approach.}
    \label{fig:evaluate_control}
\end{figure}
\subsection{Simulation experiments}
Our perception model is evaluated in closed-loop simulations, in which the control law~\eqref{eq:control_embedded} computes the commands to fulfill the desired behavior.
The evaluation considers 50 \ac{vs} task executions  performed under different initial conditions.
Each \ac{vs} task consists in driving the robot camera at a desired pose w.r.t. a box.
For each execution, the simulation environment is set up with the \ac{ooi} placed at a random pose in front of the robot, a working surface with randomized texture, and an environmental light with random orientation. 
Then, the \todo{robot camera} is moved to the desired pose using the knowledge of the simulator; at this time, the reference image $\bm{i}^\ast$ is saved, together with the \todo{object} depth to be used in the \todo{interaction matrix} computation. 
Thus, the robot is driven to a pre-defined initial posture and the VS experiment starts:
at each time step, the current camera image $\bm{i}$ is acquired to infer the features using the model $\bm{m}$; the commands are then computed using the controller~\eqref{eq:control_embedded} and sent to the robot.

To evaluate the effectiveness of our perception model in providing \todo{a} reliable feedback to the \ac{vs} law, the following ``control metrics'' are considered: (i) control effort (CE), i.e., the norm of the commands; (ii) control smoothness (CS), i.e., the norm of the commands time derivative; (iii) position error (PE), i.e.,  the Euclidean distance of the camera position from its desired value; (iv) orientation error (OE), i.e., the quaternion distance~\cite[Eq. (4)]{mahendran20173d} of the camera orientation from its desired value.
These metrics are computed only for the successful, i.e. converging, executions.
Each execution is considered successful if, in the given time of \SI{20}{s}, PE and OE fall below given threshold set to \SI{10}{cm} and 15\textdegree~(i.e. \SI{0.26}{\radian}), respectively.
CE, and CS are used to evaluate the quality of the control action produced by the neural feedback, whereas PE, OE serves as an index to establish the distance of the executions from the convergence.

The performance of our approach is compared against an end-to-end approach, that is trained to directly infer the joint velocities commands using as input camera feed and joint positions.
We consider also the joint position as input of the end-to-end to realize a fair comparison against our approach, where the joint position is indirectly considered through the Jacobian.
The end-to-end architecture shares the same layer structure as our model, with the exception of the head part, which receives the latent representation of the image concatenated with the 7 joint positions value.

\begin{table}[!t]
    \setlength{\tabcolsep}{3.5pt}
    \centering
    \caption{Median of the control metrics over the successful executions of the task in simulation
    }
    \label{tab:comparison_control}
 	\begin{tabular}{cccccc}
 	\toprule
 	Model & SR [\%] & CE [rad/s] & CS [rad/s$^2$] & PE [cm] & OE [rad] \\
 	\midrule
 	Ours &       73 & 0.37 & 15.2 & 10.8 & 0.39 \\
 	End-to-end & 38 & 0.25 & 10.0 & 13.9 & 0.32 \\
 	\bottomrule
 	\end{tabular}
\end{table}

Fig.~\ref{fig:evaluate_control} compares the values of the control metrics obtained by our model and the end-to-end approach on the successful executions of the VS task.
Table~\ref{tab:comparison_control} show the median of the metrics over the successful execution, along with the success rate (SR).
Overall we can observe a faster convergence and higher success rate of our model; we cannot see a remarkable difference in the control effort and smoothness. 

It shall be mentioned that both the \acp{nn} presented problems in correctly estimating the box orientation, accumulating a drift in the visual features measurement, in the case of our model, and in the commands in the case of the end-to-end. 
This leads to undesired effects, even after convergence, for which the controlled motion drifted. 
We overcome this issue by stopping the execution right after the satisfaction of the convergence criterion previously described.

We also carried out an ablation study of our model, and we could not observe a significant difference between the versions of the model.
\todo{This analysis suggests that there is room for improvements.}
In particular, we believe that the autoencoder could be \todo{more} beneficial for the overall approach when paired with a more advanced domain adaptation strategy~\cite{inoue2018transfer}, e.g. to cover the reality gap; instead, the state consistency loss could have a greater impact if applied to sequences longer than two samples. 

\begin{figure}[!t]
    \centering
    \includegraphics[width=0.9\columnwidth]{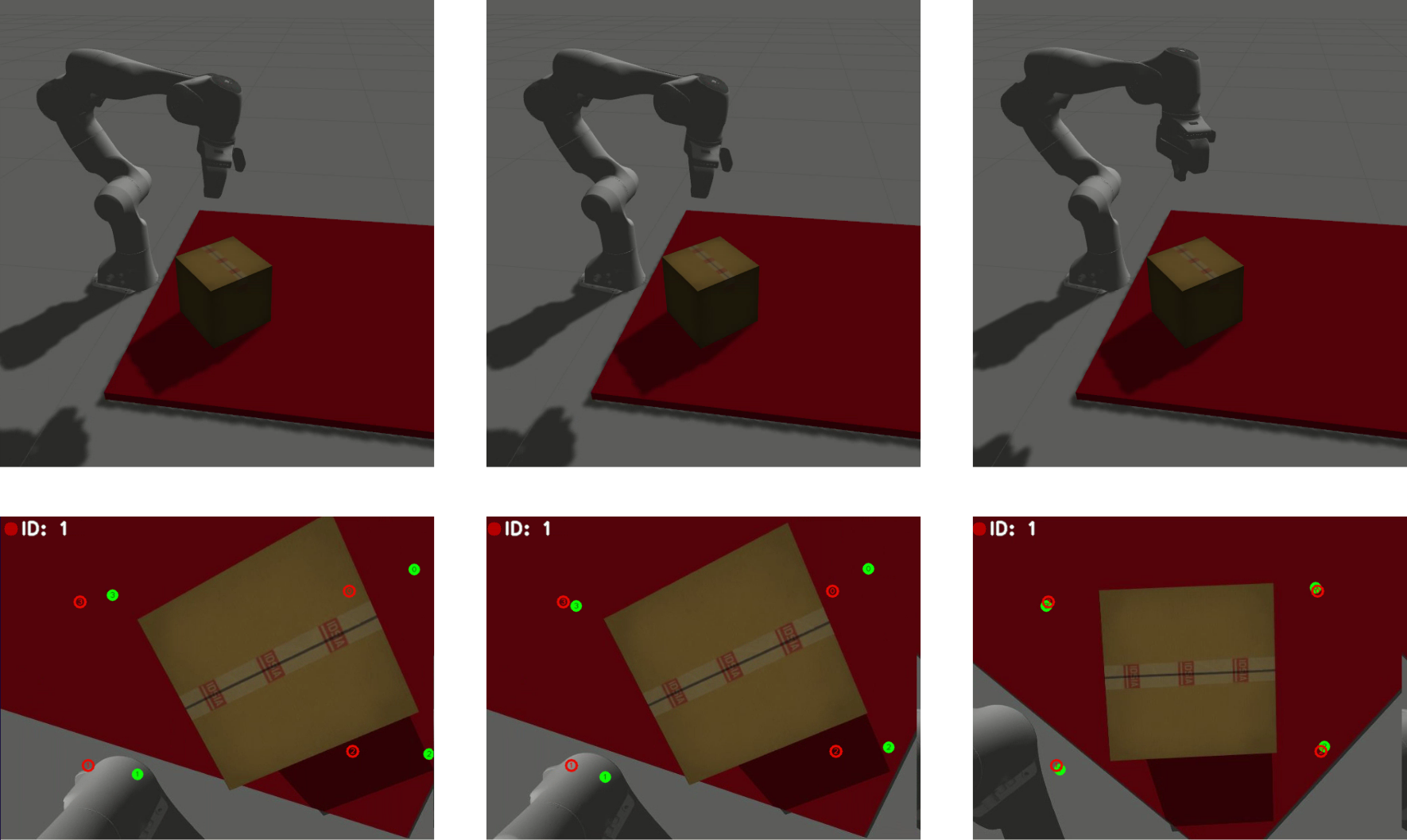}
    \caption{Execution of the VS task with the simulated robot manipulator.}
    \label{fig:sim_snap}
\end{figure}

\begin{figure}[!th]
    \centering
    \includegraphics[width=\columnwidth]{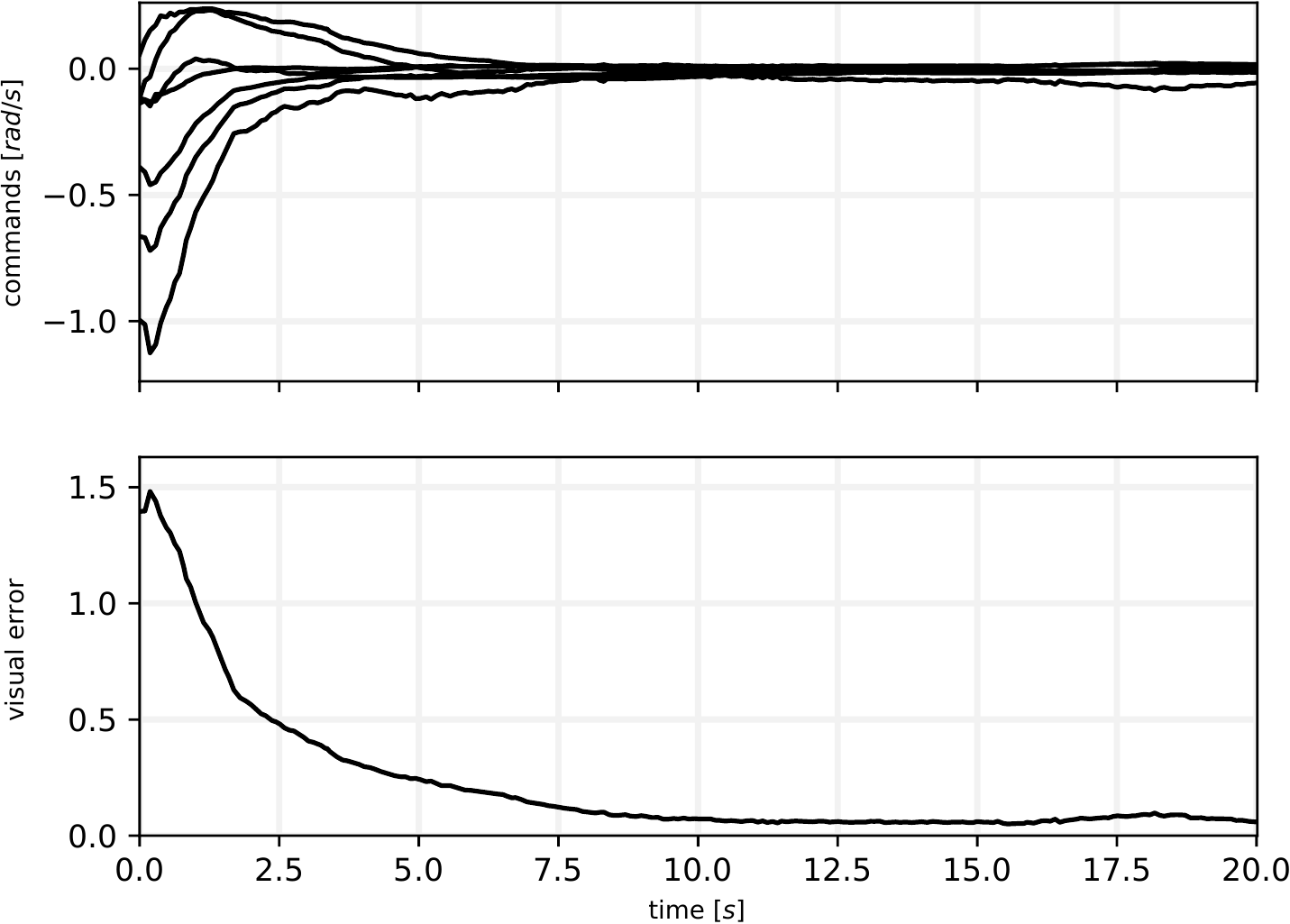}
    \caption{Control commands (top) and visual error (bottom) of the VS execution carried out with the simulated robot manipulator.}
    \label{fig:sim_plots}
\end{figure}
The execution of one single VS task carried out using our model is presented in Fig.~\ref{fig:sim_snap}. 
At the beginning of the execution, the robot camera is driven to the desired pose, the reference image taken and the corresponding desired visual features computed, by applying our model.
The desired visual features are depicted with red circles in the camera views of Fig.~\ref{fig:sim_snap}.
During the execution, the model provides the \ac{vs} with current visual features, depicted with green dots, till converge, i.e. when the camera pose value goes under a certain threshold.
Fig.~\ref{fig:sim_plots} shows the computed commands and the norm of the visual error for this simulation execution.

\subsection{Real experiments}
We consider our model also for closed-loop experiments carried out with the real robot manipulator. 
\todo{To this end,} we considered the model trained only on randomized simulated data, and trained with additional data augmentations designed to replicate noise and interference present in the real camera sensor.
Given the low performance of the model in estimating an accurate orientation of the \ac{ooi}, we decided to change the structure of the control \todo{considering only the position. In practice, we projected} the~\ac{vs} task into the null space of another primary task, consisting in keeping a constant camera orientation.
Crucially, this design change in the control law was possible
thanks to the flexibility of our approach, where the perception model is decoupled from the controller. 
Indeed, even if our model uses the knowledge of the classical \ac{vs} scheme during training, its output can be used for deployment in other contexts; whereas the end-to-end approach does not offer such flexibility.

\begin{figure}[!t]
    \centering
    \includegraphics[width=0.9\columnwidth]{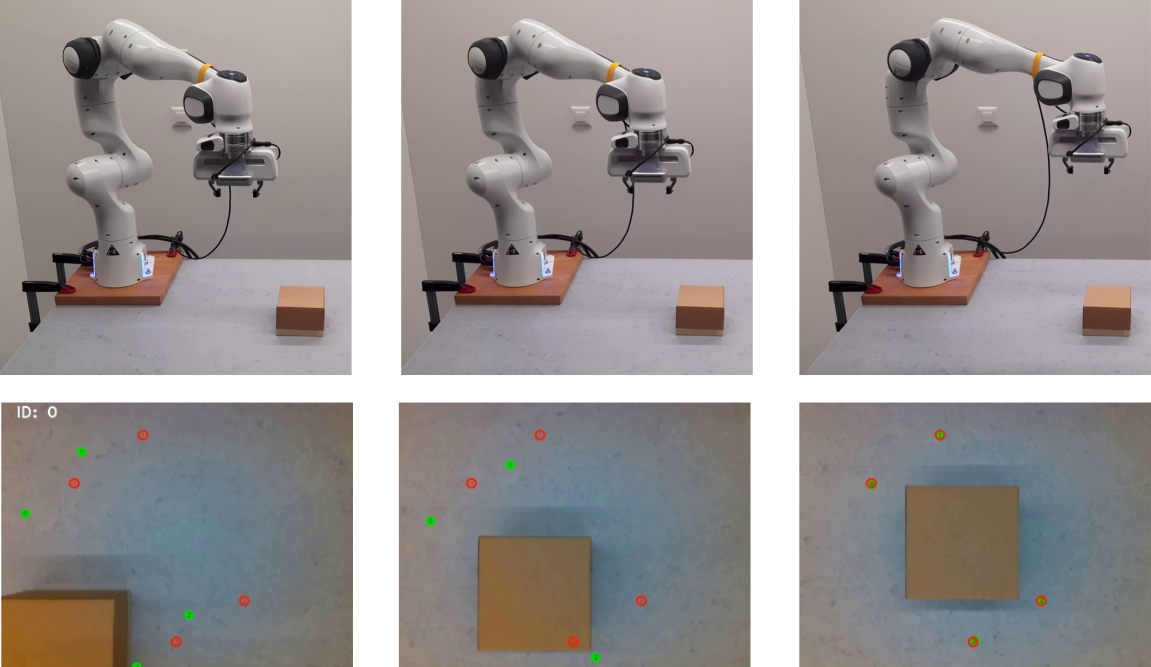}
    \caption{Execution of the VS task with the real robot manipulator.
    }
    \label{fig:exp}
\end{figure}

In Fig.~\ref{fig:exp} we show three snapshots of the robot performing the control task, along with the corresponding camera views.
The experiment starts with the \ac{ooi} partially out of the camera's field of view -- a challenging situation that could result in a failed attempt when using explicit methods for the image processing.
Our neural perception model can provide the required feedback for the correct convergence of the \ac{vs} task.
A video of this experiment, as well as other executions and the simulations, are shown in the multimedia attachment.

\section{Conclusion and Future Work}\label{sec:conclusion}

We have presented an approach for learning a deep perception model, providing feedback to \todo{visual controllers}. 
The neural perception model is trained \todo{leveraging} the knowledge of the \todo{visual servoing} and the features' \todo{model} as supervision.
As a result, we could estimate geometrically interpretable neural visual features, so that they could be used in the analytical form of the visual servoing.
\todo{By keeping perception algorithm and controller decoupled}, we could preserve flexibility \todo{and modularity} of the framework.
In fact, \todo{our} neural perception could be used in different control structures.

\todo{We have shown promising results,} and future work will be devoted to achieving higher performances.
To this end, it might be necessary to sophisticate and redesign the architecture of our \ac{nn}.
\todo{Further development will be done to emphasize the contribution of the different loss terms. In fact, we believe that by implementing a different strategy of data shuffling, we could obtain higher performance of the state consistency loss.}
In the future, we will also tackle the sim-to-real gap, by exploiting domain adaptation techniques that can be easily implemented thanks to the auto-encoding part of our model.
\todo{These aspects will help on improving the estimation of the tracked object orientation.}
Bayesian optimization techniques can be used to optimally weigh the different loss functions during the training.
Also, the perception model could be trained to optimize the task performance, so that the estimated feedback could be even more reliable for the downstream visual controller.
\todo{An obvious improvement could also come from more informative training data, for which we will consider the use of a planner.}
\todo{Furthermore,} we believe that the model could benefit from other sources of sensory information, that can be handled by our network design.
\todo{Finally, we plan to carry out more experiments, training our perception model with different type and number of features, and test it in different control contexts and with more complex target objects.}

\addtolength{\textheight}{-1.50cm} 

\bibliographystyle{IEEEtran}
\bibliography{biblio.bib}

\end{document}